\pdfoutput=1

\documentclass{article}
\usepackage[T1]{fontenc}
\usepackage[utf8]{inputenc}
\usepackage{ismir}
\usepackage{amsmath,amssymb,cite,url}
\usepackage{graphicx}
\usepackage{color}
\usepackage{booktabs}
\usepackage{multirow}
\usepackage{subcaption}

\renewcommand{\permission}{%
  \begin{figure}[b]%
    \footnotesize\noindent\rule{0.4\columnwidth}{0.4pt}\par\vspace{2pt}%
    \noindent$^{*}$Corresponding author.%
  \end{figure}}

\newcommand{\ours}{\textsc{MuQ-Eval}}

\title{MuQ-Eval: An Open-Source Per-Sample Quality Metric for AI Music Generation Evaluation}

\threeauthors
  {Di Zhu} {\normalsize Stevens Institute of Technology \\ \normalsize Hoboken, NJ, USA \\ \normalsize\texttt{dzhu1@stevens.edu}}
  {Zihan Chen$^{*}$} {\normalsize Stevens Institute of Technology \\ \normalsize Hoboken, NJ, USA \\ \normalsize\texttt{zchen61@stevens.edu}}
  {Zixuan Li} {\normalsize Columbia University \\ \normalsize New York, NY, USA \\ \normalsize\texttt{zixuan.li@columbia.edu}}

\def\authorname{D. Zhu, Z. Chen, and Z. Li}

\begin{document}

\maketitle

\begin{abstract}
Distributional metrics such as Fr\'{e}chet Audio Distance cannot score individual music clips and correlate poorly with human judgments, while the only per-sample learned metric achieving high human correlation is closed-source. We introduce \ours{}, an open-source per-sample quality metric for AI-generated music built by training lightweight prediction heads on frozen MuQ-310M features using MusicEval, a dataset of generated clips from 31 text-to-music systems with expert quality ratings. Our simplest model, frozen features with attention pooling and a two-layer MLP, achieves system-level SRCC = 0.957 and utterance-level SRCC = 0.838 with human mean opinion scores. A systematic ablation shows that no addition meaningfully improves the frozen baseline, indicating that frozen MuQ representations already capture quality-relevant information. Encoder choice is the dominant design factor. LoRA-adapted models trained on as few as 150 clips already achieve usable correlation, enabling personalized quality evaluators from individual listener annotations. A controlled degradation analysis reveals selective sensitivity to signal-level artifacts but insensitivity to musical-structural distortions.
\end{abstract}

\section{Introduction}

Text-to-music (TTM) generation has advanced rapidly. Systems such as MusicGen~\cite{copet2023musicgen}, MusicLM~\cite{agostinelli2023musiclm}, and Stable Audio~\cite{evans2024stableaudio} can now produce multi-instrument arrangements from short text prompts. Yet this progress has exposed a critical bottleneck: \emph{the field lacks reliable, automatic methods for assessing the perceptual quality of generated music}. Without such methods, researchers cannot rigorously compare systems, practitioners cannot filter low-quality outputs at scale, and reinforcement-from-feedback training loops lack a trustworthy reward signal.

Music quality is a multi-dimensional percept encompassing timbral naturalness, harmonic coherence, rhythmic stability, and production polish, all of which are difficult to define formally and expensive to annotate. Human listening studies remain the gold standard, but they are slow, costly, and poorly reproducible across labs.

Existing automatic metrics fail to fill this gap. Distributional metrics such as Fr\'{e}chet Audio Distance (FAD)~\cite{kilgour2019fad} cannot score individual clips and correlate weakly with human preferences ($\tau = 0.14$ for VGGish~\cite{huang2025mad}). Text-audio alignment scores such as CLAP~\cite{wu2023clap, elizalde2024clap} measure semantic relevance rather than perceptual quality. The per-sample learned metrics that do exist are either weakly correlated (Audiobox Aesthetics: $r = 0.200$~\cite{zhang2025aesthetics}) or closed-source (DORA-MOS: SRCC $= 0.988$~\cite{audiomos2025}).

This stands in sharp contrast to adjacent domains where deep features combined with quality annotations have produced highly effective metrics: LPIPS~\cite{zhang2018lpips} in vision (68.9\% 2AFC agreement), NISQA~\cite{mittag2021nisqa} and DNSMOS~\cite{reddy2022dnsmos} in speech (PCC up to 0.98). The AudioMOS 2025 Challenge~\cite{audiomos2025} showed this recipe works for music, but winning systems remain closed-source.

This paper asks: \emph{can the ``deep features + quality annotations'' recipe produce an open per-sample quality metric for generated music, and which pipeline components actually matter?} We investigate through a progressive ablation on MusicEval~\cite{musiceval2025}, yielding four principal findings:

\begin{enumerate}
    \item \textbf{Frozen MuQ features are remarkably effective.} A simple baseline consisting of frozen MuQ-310M~\cite{zhu2025muq} features with attention pooling and a two-layer MLP achieves system-level SRCC $= 0.957$ and utterance-level SRCC $= 0.838$.

    \item \textbf{Progressive training complexity does not help.} No ablation component (ordinal loss, LoRA, contrastive loss, uncertainty weighting) improves system-level SRCC by $\geq 0.02$.

    \item \textbf{Encoder choice is the dominant factor.} Replacing MuQ-310M with MERT-95M~\cite{yuan2024mert} reduces system-level SRCC from 0.957 to 0.946.

    \item \textbf{Remarkably few annotations are needed.} LoRA-adapted models reach utterance-level SRCC $= 0.761$ with only 150 training clips, making personalized quality evaluators practical from small annotation budgets.
\end{enumerate}

We release \ours{} as a fully open-source metric including model weights, training code, and evaluation benchmark.\footnote{Code and model weights will be released upon acceptance.} The pipeline trains in under 2~GPU-hours on a single NVIDIA RTX~4080.

The remainder of this paper is organized as follows. Section~\ref{sec:related} surveys distributional, alignment, and per-sample learned metrics. Section~\ref{sec:method} describes the encoder-pooling-head architecture and progressive training variants. Section~\ref{sec:experiments} details the experimental setup and evaluation protocol. Section~\ref{sec:results} presents system-level and utterance-level results, the ablation study, degradation analysis, and data efficiency experiments. Section~\ref{sec:discussion} interprets the findings and discusses limitations. Section~\ref{sec:conclusion} concludes.

\section{Related Work}
\label{sec:related}

Kader and Karmaker~\cite{kader2025survey} survey evaluation metrics for music generation, and Vinay and Lerch~\cite{vinay2022evaluating} show in a listening study that commonly used objective metrics do not adequately describe the perceptual quality of generative audio systems. We focus here on work that bears directly on per-sample quality prediction.

\subsection{Distributional and Alignment Metrics}

FAD~\cite{kilgour2019fad} measures the distance between Gaussians fitted to reference and generated embeddings. It therefore scores a set of clips rather than an individual clip, and its agreement with listeners depends on the embedding: $\tau = 0.14$ with VGGish on text-to-music preferences~\cite{huang2025mad}, but $\rho > 0.5$ with PANNs on environmental audio~\cite{tailleur2024fad}. Variants adapt FAD to music~\cite{gui2024fad}, replace the Gaussian assumption with a kernel-based maximum mean discrepancy (KAD)~\cite{chung2025kad}, or, as in the MAUVE Audio Divergence (MAD)~\cite{huang2025mad}, reach $\tau = 0.62$ with self-supervised music embeddings. Gr\"{o}tschla et al.~\cite{grotschla2025} compare such metrics with a large human preference study and find that every tested metric misranks at least one system. CLAP score~\cite{wu2023clap, elizalde2024clap} measures text--audio semantic agreement rather than quality, and Human-CLAP~\cite{takano2025humanclap} fine-tunes CLAP on human relevance judgments to improve that agreement. Neither family yields a per-clip estimate of perceptual quality.

\subsection{Learned Quality Prediction in Speech and Vision}

Per-sample learned metrics are well established in neighboring fields. In vision, LPIPS~\cite{zhang2018lpips} reaches 68.9\% 2AFC agreement with human similarity judgments using linear probes on deep features. In speech, MOS prediction moved from networks trained from scratch, such as MOSNet~\cite{lo2019mosnet}, to pretrained self-supervised encoders with lightweight output layers. Cooper et al.~\cite{cooper2022generalization} found that fine-tuned wav2vec~2.0 models generalize well to out-of-domain listening tests, while unseen systems remain especially hard to predict; top systems in the VoiceMOS Challenge 2022~\cite{huang2022voicemos}, such as UTMOS~\cite{saeki2022utmos}, likewise build on self-supervised features. NISQA~\cite{mittag2021nisqa} (PCC $= 0.80$--$0.95$) and DNSMOS~\cite{reddy2022dnsmos} (PCC up to $0.98$) are widely used non-intrusive predictors, SCOREQ~\cite{ragano2024scoreq} improves domain generalization with contrastive regression, and ALLD~\cite{chen2025alld} turns audio large language models into descriptive quality evaluators. \ours{} asks whether the same ``pretrained encoder plus light head'' recipe carries over to generated music.

\subsection{Learned Quality Prediction for Music}

For music and general audio, PAM~\cite{deshmukh2024pam} prompts a frozen audio--language model without quality-specific training, and Meta Audiobox Aesthetics~\cite{tjandra2025audiobox} predicts four aesthetic axes (production quality, production complexity, content enjoyment, and content usefulness) for speech, music, and sound. Its predictions correlate only weakly with text-to-music preferences ($r = 0.200$~\cite{zhang2025aesthetics}), and Li et al.~\cite{li2026meva} report correlations of about $0.2$ when pooling five music rating benchmarks; they show instead that internal signals of a frozen MusicGen model (token loss, predictive entropy, and sparse-autoencoder concepts) predict human ratings well.

The AudioMOS Challenge 2025~\cite{audiomos2025} adopted MusicEval for Track~1, with a CLAP-based baseline that predicts musical impression (MI) and textual alignment (TA) through two MLP heads. The winning ASTAR-NTU submission~\cite{ritter2025astar} introduced DORA-MOS: frozen MuQ features pass through a Transformer layer, attention pooling, and an MLP to predict MI, while a cross-attention branch over MuQ and RoBERTa text features predicts TA. Scores are learned as Gaussian-softened distributions over 20 bins, and the final submission stacks nine models with ridge regression, reaching system-level SRCC $= 0.988$ for MI and $0.944$ for TA on the official test set. The team also tried CLAP, MERT, and MuQ-MuLan audio encoders and found MuQ best. The MI branch of DORA-MOS is the closest prior design to ours: \ours{} drops the Transformer layer, the text branch, and the ensemble, and to our knowledge no code or weights accompany the DORA-MOS system description. QAMRO~\cite{wang2025qamro} keeps the CLAP baseline architecture and adds a quality-aware adaptive-margin ranking loss, reaching system-level MI SRCC $= 0.972$ on the challenge test set, with code released.

Song-level evaluation has moved in the same direction. SongEval~\cite{songeval2025} provides an evaluation toolkit alongside its ratings. Lv et al.~\cite{lv2026song} fuse MuQ features of the full mix with those of separated vocal and accompaniment stems and predict scores through hierarchical interval classification. SongBench~\cite{wu2026songbench} trains a MuQ-based evaluator on seven expert-rated aspects of full-length songs and tests it on generators held out from training. MuQ has thus become a common backbone for music quality prediction, in line with our encoder comparison (Section~\ref{sec:results}).

Per-sample predictors also serve as training signals. MusicRL~\cite{cideron2024musicrl} fine-tunes MusicLM with reinforcement learning from rater-derived rewards and large-scale user preferences, and SMART~\cite{jonason2025smart} optimizes a symbolic piano generator with GRPO, using the Audiobox Aesthetics content-enjoyment score of rendered 10-second excerpts as the reward. Once a metric becomes a reward, its blind spots become optimization targets, which makes open and well-characterized metrics more important.

\subsection{Datasets and Encoders}

MusicEval~\cite{musiceval2025} provides 2,748 clips from 31 TTM systems with 13,740 expert ratings of MI and TA; AudioMOS 2025 split it 70/15/15 into 1,923 training, 412 development, and 413 test clips~\cite{audiomos2025}. SongEval~\cite{songeval2025} contains 2,399 full-length songs (about 140 hours) rated by 16 professional annotators on five aesthetic dimensions ($\sim$48k ratings), and SongBench~\cite{wu2026songbench} contains 11,717 songs rated on seven aspects. Preference resources include MusicPrefs~\cite{huang2025mad}, the human study of Gr\"{o}tschla et al.~\cite{grotschla2025}, and Music Arena~\cite{kim2025musicarena}, a live platform that collects pairwise preferences between TTM systems on user-written prompts and releases them on a rolling basis.

MERT~\cite{yuan2024mert} is a music self-supervised model trained with acoustic (RVQ-VAE) and musical (CQT) teacher targets; its largest variant is pre-trained on 160K hours. MuQ~\cite{zhu2025muq} instead predicts Mel residual vector quantization (Mel-RVQ) tokens; trained on only 0.9K hours of open data it already outperforms earlier music self-supervised models, and its iteratively refined variant reaches an average MARBLE~\cite{yuan2023marble} score of 77.0 versus 74.4 for MERT. MuQ-MuLan extends MuQ to a joint music--text embedding space~\cite{zhu2025muq}. We use the public MuQ checkpoint trained on the Million Song Dataset (MuQ-large-msd-iter).

\section{Method}
\label{sec:method}

\subsection{Architecture}

\ours{} follows a simple encoder-pooling-head architecture. Given a music audio waveform $\mathbf{x} \in \mathbb{R}^{T}$ at 24~kHz, we extract frame-level features $\mathbf{H} = [\mathbf{h}_1, \ldots, \mathbf{h}_L] \in \mathbb{R}^{L \times d}$ from a pre-trained encoder.

\subsubsection{Encoder}
We use MuQ-310M~\cite{zhu2025muq}, a Wav2Vec2-Conformer model with 24 layers and $d = 1024$, pre-trained for music understanding at 24~kHz with 50~Hz frame rate.

\subsubsection{Pooling}
Frame-level features are aggregated via learned attention pooling:
\begin{equation}
    \mathbf{z} = \sum_{l=1}^{L} \alpha_l \mathbf{h}_l, \quad \alpha_l = \frac{\exp(\mathbf{w}^\top \mathbf{h}_l)}{\sum_{l'} \exp(\mathbf{w}^\top \mathbf{h}_{l'})}
\end{equation}
where $\mathbf{w} \in \mathbb{R}^d$ is a learnable attention vector.

\subsubsection{Prediction heads}
The pooled representation $\mathbf{z}$ is passed to separate 2-layer MLP heads (GELU, hidden dim 256) for each quality dimension:
\begin{equation}
    \hat{y} = W_2 \cdot \text{GELU}(W_1 \mathbf{z} + b_1) + b_2
\end{equation}

\subsection{Training Variants}

We investigate a progressive sequence of training configurations:

\subsubsection{A1: Frozen encoder + MSE (baseline)}
Only attention pooling and MLP heads are trained ($\sim$1M parameters) with MSE loss:
\begin{equation}
    \mathcal{L}_\text{MSE} = \frac{1}{N}\sum_i (\hat{y}_i - y_i)^2
\end{equation}

\subsubsection{A2: Frozen encoder + ordinal classification}
Following DORA-MOS~\cite{audiomos2025}, we use Gaussian-softened ordinal cross-entropy with $K = 5$ bins:
\begin{equation}
    p_k(y) \propto \exp\left(-\frac{(y - c_k)^2}{2\sigma^2}\right), \quad \mathcal{L}_\text{ord} = -\sum_k p_k(y) \log \hat{p}_k
\end{equation}
where $c_k$ is the bin center and $\sigma = 0.5$.

\subsubsection{A3a: LoRA encoder adaptation}
We add LoRA~\cite{hu2022lora} adapters to all attention projections (rank $r = 16$, $\alpha = 32$), adding $\sim$2M trainable parameters.

\subsubsection{A3b: + Contrastive auxiliary loss}
A pairwise contrastive loss for pairs with $y_i > y_j + m$ ($m = 0.5$):
\begin{equation}
    \mathcal{L} = \mathcal{L}_\text{ord} + 0.5 \cdot \max(0, -(\hat{y}_i - \hat{y}_j) + m)
\end{equation}
with the contrastive term warm-started at epoch 6.

\subsubsection{A3c: + Uncertainty weighting}
Following Kendall et al.~\cite{kendall2018uncertainty}, each head learns a log-variance scaling its loss:
\begin{equation}
    \mathcal{L}_\text{total} = \sum_h \frac{1}{2\sigma_h^2} \mathcal{L}_h + \log \sigma_h
\end{equation}

\subsubsection{A4: MERT-95M encoder ablation}
We replace MuQ-310M with MERT-95M~\cite{yuan2024mert} using full fine-tuning with ordinal loss to isolate encoder choice.

\section{Experimental Setup}
\label{sec:experiments}

\subsection{Dataset}

We train and evaluate on MusicEval~\cite{musiceval2025}: 2,748 clips from 31 TTM systems with expert ratings on musical impression (MI) and textual alignment (TA) on a 1--5 Likert scale (5 ratings/clip, 13,740 total). Under 5-fold CV, each fold uses ${\sim}1{,}540$ clips for training and ${\sim}385$ for testing.

\subsection{Evaluation Protocol}

We use 5-fold cross-validation stratified by TTM model. \textbf{System-level:} mean predicted and human MOS per model, then SRCC, PCC, and $\tau$ across 31 models. \textbf{Utterance-level:} per-clip PCC and SRCC on each fold's test set. All metrics are 5-fold means $\pm$ SD with 95\% BCa bootstrap CIs ($B = 1000$). Pairwise comparisons use the Steiger test with Bonferroni correction ($\alpha_\text{adj} = 0.01$); ablation deltas assessed against $\Delta \geq 0.02$.

\subsection{Baselines}

FAD (VGGish) is computed on each fold's test set as a distributional baseline. Audiobox Aesthetics correlation ($r = 0.200$) is referenced from Zhang et al.~\cite{zhang2025aesthetics}.

\subsection{Training Details}

\begin{figure}[t]
    \centering
    \includegraphics[alt={Training curves showing validation SRCC and training loss across epochs for all model configurations on fold 0},width=\columnwidth]{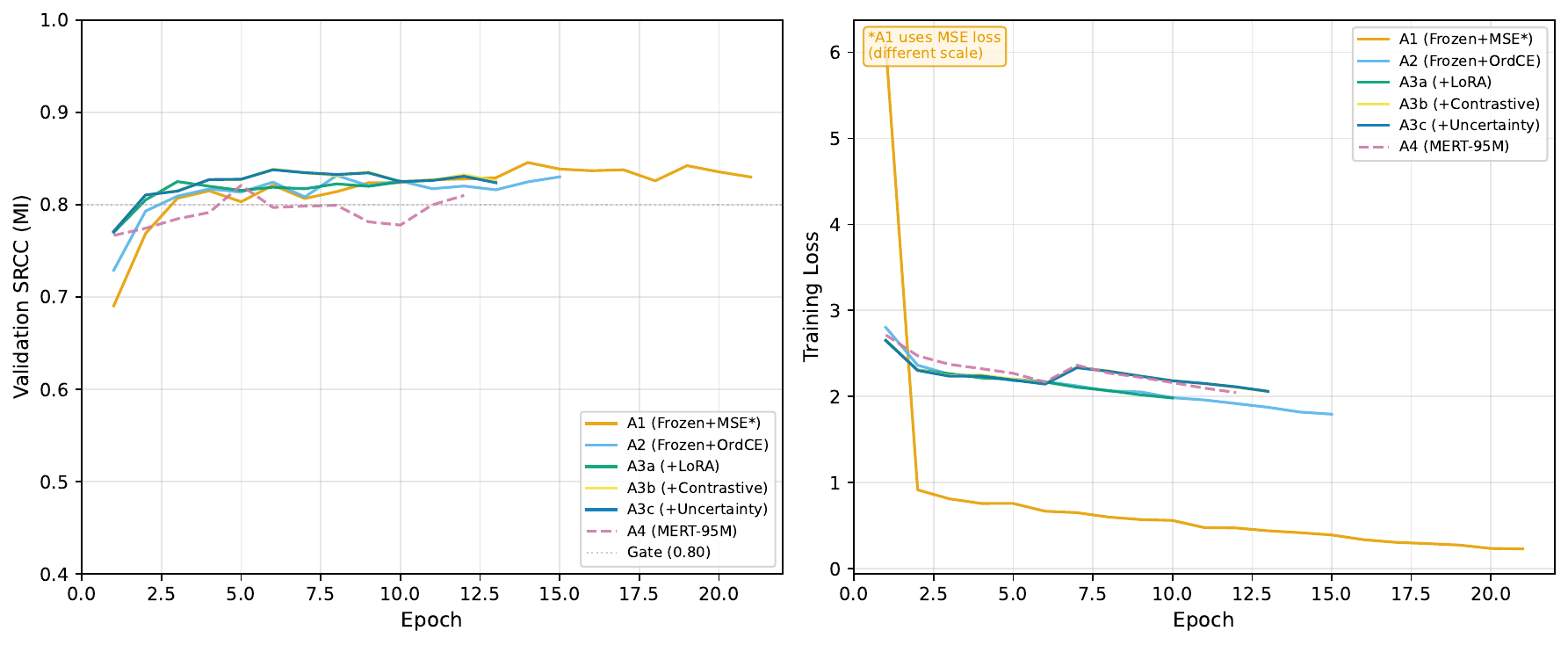}
    \caption{Training dynamics: validation SRCC and training loss for all configurations (fold~0). All MuQ variants converge to similar validation SRCC.}
    \label{fig:training_curves}
\end{figure}

All models: AdamW ($3 \times 10^{-4}$ for heads, $1 \times 10^{-5}$ for LoRA), batch size 16, up to 50 epochs with early stopping (patience 10). Mixed precision (bf16) on a single NVIDIA RTX 4080 (16~GB). Audio: 24~kHz, 10-second clips. Training curves are shown in \figref{fig:training_curves}.

\section{Results}
\label{sec:results}

\subsection{Main Results: System-Level Correlation}

\tabref{tab:system_level} presents system-level correlations. All MuQ variants achieve SRCC $> 0.95$, substantially exceeding the 0.90 target and Audiobox Aesthetics ($r = 0.200$).

\begin{table}[t]
\centering
\caption{System-level correlation with human MOS (31 TTM models, 5-fold CV). Best in \textbf{bold}; 95\% BCa bootstrap CIs in brackets.}
\label{tab:system_level}
\small
\begin{tabular}{@{}lccc@{}}
\toprule
Method & SRCC$_\text{sys}$ & PCC$_\text{sys}$ & $\tau_\text{sys}$ \\
\midrule
FAD (VGGish)$^\dagger$ & 0.295{\scriptsize$\pm$.031}$^\ddagger$ & --- & 0.14 \\
Audiobox Aes.$^*$ & 0.200 & --- & --- \\
\midrule
A1 (Frozen+MSE) & 0.957{\scriptsize$\pm$.014} & 0.960{\scriptsize$\pm$.013} & 0.839{\scriptsize$\pm$.019} \\
 & {\scriptsize[.898, .986]} & {\scriptsize[.916, .980]} & {\scriptsize[.716, .910]} \\
A2 (Frozen+Ord.) & 0.953{\scriptsize$\pm$.015} & 0.963{\scriptsize$\pm$.007} & 0.839{\scriptsize$\pm$.034} \\
A3a (+LoRA) & \textbf{0.960}{\scriptsize$\pm$.017} & \textbf{0.969}{\scriptsize$\pm$.009} & \textbf{0.846}{\scriptsize$\pm$.034} \\
A3b (+Contr.) & 0.955{\scriptsize$\pm$.016} & 0.967{\scriptsize$\pm$.010} & 0.838{\scriptsize$\pm$.035} \\
A3c (+Unc.Wt.) & 0.955{\scriptsize$\pm$.016} & 0.968{\scriptsize$\pm$.010} & 0.836{\scriptsize$\pm$.035} \\
A4 (MERT-95M) & 0.946{\scriptsize$\pm$.012} & 0.948{\scriptsize$\pm$.013} & 0.822{\scriptsize$\pm$.022} \\
\bottomrule
\multicolumn{4}{@{}l@{}}{\scriptsize $^\dagger$Published $\tau$ from~\cite{huang2025mad}; $^\ddagger$raw FAD (lower is better); $^*$from~\cite{zhang2025aesthetics}.}
\end{tabular}
\end{table}

\begin{figure}[t]
    \centering
    \includegraphics[alt={Scatter plot of predicted MuQ-Eval scores versus human MOS for 31 text-to-music models showing strong linear correlation},width=\columnwidth]{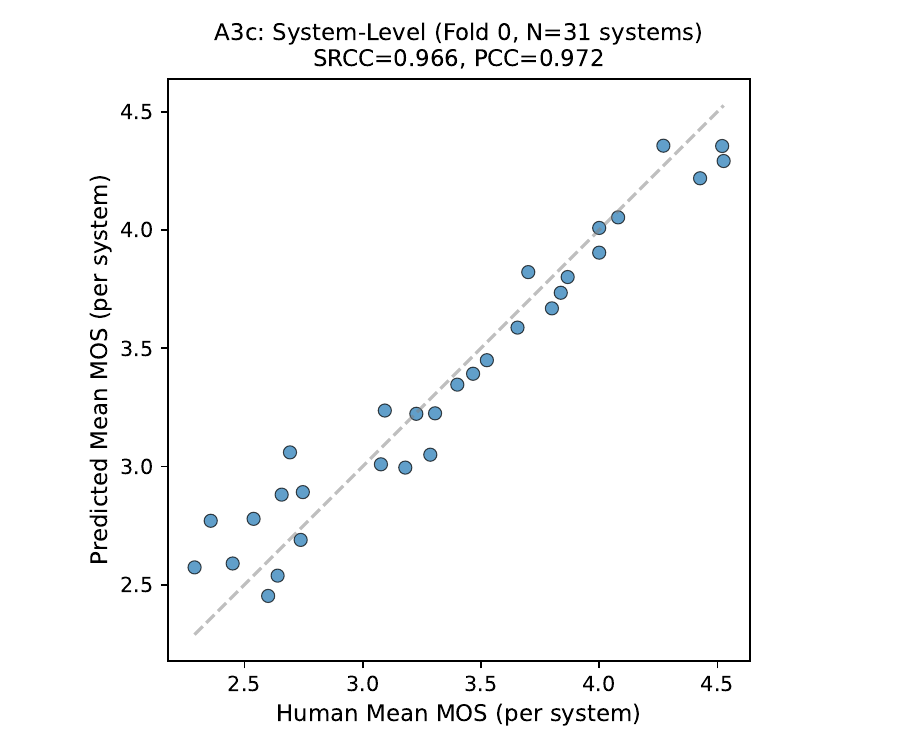}
    \caption{System-level scatter: predicted \ours{} scores vs.\ human MOS for 31 TTM models (A1, fold~0).}
    \label{fig:system_scatter}
\end{figure}

The simplest model A1 achieves SRCC $= 0.957$, nearly matching A3a (0.960). The Steiger test confirms no significant difference between A1 and any other MuQ variant ($p > 0.01$). \figref{fig:system_scatter} visualizes the system-level correspondence.

\subsection{Utterance-Level Correlation}

\begin{figure}[t]
    \centering
    \includegraphics[alt={Scatter plot of predicted scores versus human MOS at the utterance level showing strong per-clip correlation with most predictions within half a point of ground truth},width=\columnwidth]{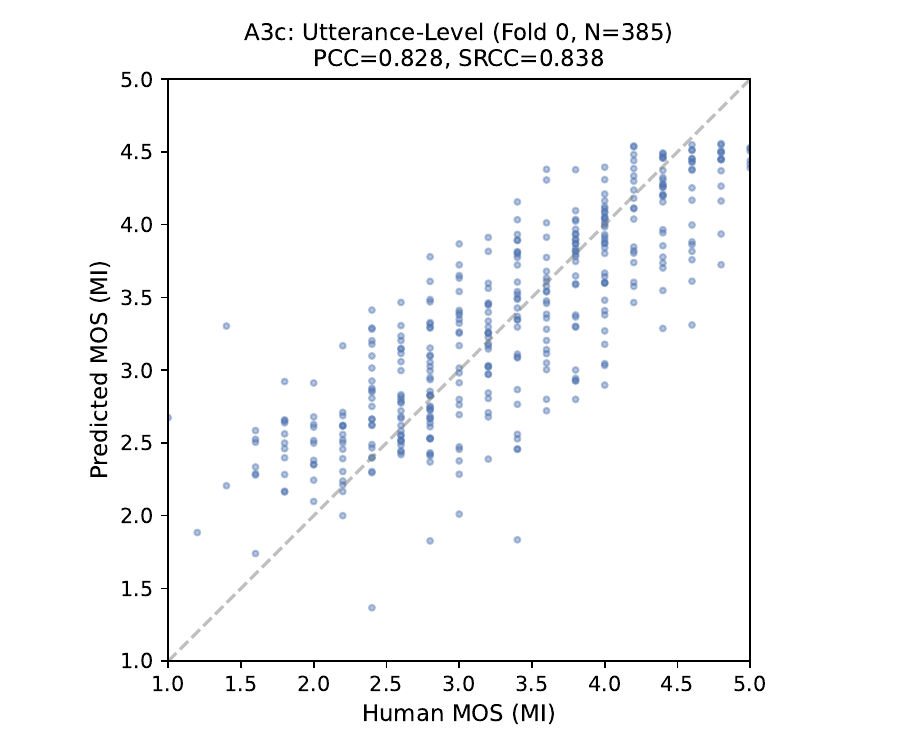}
    \caption{Utterance-level scatter: predicted \ours{} scores vs.\ human MOS for individual clips (A1, fold~0, MI dimension). The diagonal represents perfect prediction.}
    \label{fig:utterance_scatter}
\end{figure}

\tabref{tab:utterance_level} shows per-clip correlations. All models achieve MI SRCC $> 0.82$, exceeding the 0.65 target. The MI head substantially outperforms the TA head (SRCC 0.83--0.84 vs.\ 0.58--0.62), suggesting TA prediction requires text-conditioned architectures. \figref{fig:utterance_scatter} visualizes the per-clip correspondence for A1, showing that most predictions fall within $\pm 0.5$ of the human MOS.

\begin{table}[t]
\centering
\caption{Utterance-level correlation with human MOS (MI dimension, $\sim$385 clips/fold, 5-fold CV).}
\label{tab:utterance_level}
\small
\begin{tabular}{@{}lcccc@{}}
\toprule
Method & MI PCC & MI SRCC & TA PCC & TA SRCC \\
\midrule
A1 & .828{\scriptsize$\pm$.017} & .838{\scriptsize$\pm$.016} & .594{\scriptsize$\pm$.041} & .591{\scriptsize$\pm$.034} \\
A2 & .824{\scriptsize$\pm$.020} & .833{\scriptsize$\pm$.019} & .604{\scriptsize$\pm$.024} & .601{\scriptsize$\pm$.023} \\
A3a & .827{\scriptsize$\pm$.020} & .835{\scriptsize$\pm$.019} & \textbf{.614}{\scriptsize$\pm$.029} & \textbf{.614}{\scriptsize$\pm$.029} \\
A3b & \textbf{.829}{\scriptsize$\pm$.019} & \textbf{.840}{\scriptsize$\pm$.019} & .588{\scriptsize$\pm$.036} & .585{\scriptsize$\pm$.034} \\
A3c & \textbf{.829}{\scriptsize$\pm$.019} & \textbf{.840}{\scriptsize$\pm$.019} & .587{\scriptsize$\pm$.038} & .584{\scriptsize$\pm$.035} \\
A4 & .808{\scriptsize$\pm$.019} & .825{\scriptsize$\pm$.015} & .613{\scriptsize$\pm$.034} & .619{\scriptsize$\pm$.028} \\
\bottomrule
\end{tabular}
\end{table}

\subsection{Ablation Analysis}

\tabref{tab:ablation} presents the progressive ablation results. No component achieves the pre-registered $\Delta \geq 0.02$ threshold at system level.

\textbf{Ordinal CE vs.\ MSE} ($\Delta = -0.004$). Gaussian-softened ordinal classification, used successfully in DORA-MOS~\cite{audiomos2025}, does not improve over simple MSE regression. Cohen's $q = 0.050$ indicates a negligible effect. At utterance level, ordinal CE shows a small improvement in TA correlation ($+0.010$ SRCC) but this is offset by a decrease in MI correlation ($-0.005$ SRCC).

\textbf{LoRA adaptation} ($\Delta = +0.007$). Fine-tuning the encoder with LoRA produces the largest positive delta but remains below threshold. The effect is small (Cohen's $q = 0.085$) and adds 2M trainable parameters and training complexity.

\textbf{Contrastive loss} ($\Delta = -0.004$). Adding the pairwise contrastive objective slightly \emph{degrades} system-level performance. At utterance level, it creates a trade-off: MI SRCC increases by $+0.005$ (A3b vs.\ A3a) but TA SRCC decreases by $-0.029$, suggesting the contrastive signal conflicts with the textual alignment task.

\textbf{Uncertainty weighting} ($\Delta = -0.000$). Learned uncertainty weighting has no measurable effect, with Cohen's $q = 0.003$, functionally zero.

\begin{figure}[t]
    \centering
    \includegraphics[alt={Bar chart comparing system-level and utterance-level SRCC across all ablation experiments showing tight clustering above 0.90},width=\columnwidth]{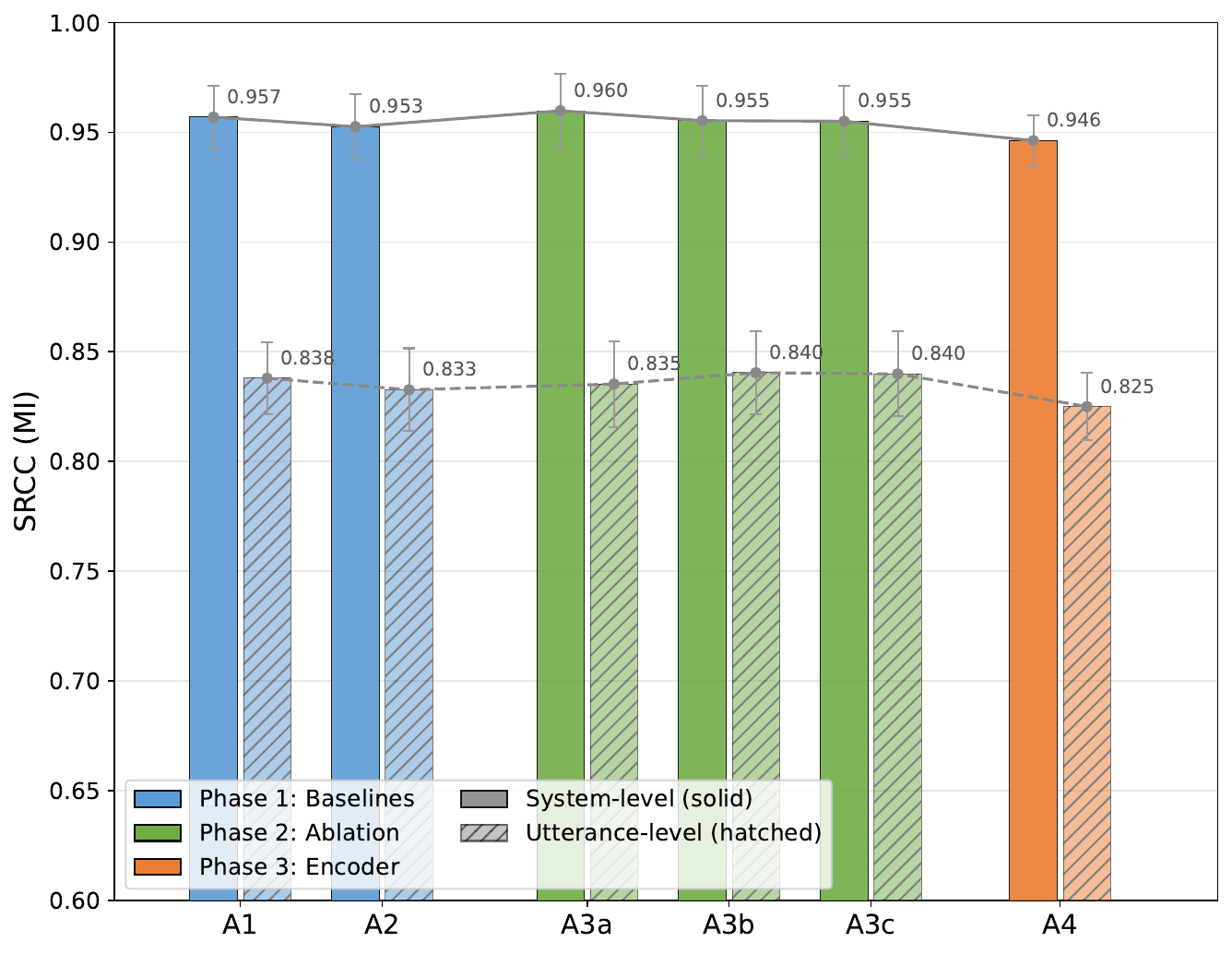}
    \caption{System-level and utterance-level SRCC(MI) per experiment. All MuQ variants cluster tightly, confirming progressive complexity does not yield improvement.}
    \label{fig:ablation_bar}
\end{figure}

\figref{fig:ablation_bar} confirms all MuQ variants cluster tightly at both system and utterance levels.

\begin{table}[t]
\centering
\caption{Progressive ablation deltas (system-level SRCC). Threshold: $\Delta \geq 0.02$. None met.}
\label{tab:ablation}
\small
\begin{tabular}{@{}llccc@{}}
\toprule
Step & Component & $\Delta$ SRCC$_\text{sys}$ & Cohen's $q$ & $\geq 0.02$? \\
\midrule
A2$-$A1 & Ordinal CE & $-0.004$ & 0.050 & No \\
A3a$-$A2 & LoRA tuning & $+0.007$ & 0.085 & No \\
A3b$-$A3a & Contrastive & $-0.004$ & 0.054 & No \\
A3c$-$A3b & Unc.\ weight & $-0.000$ & 0.003 & No \\
\bottomrule
\end{tabular}
\end{table}

\subsection{Encoder Comparison}

Replacing MuQ-310M with MERT-95M (A4) yields the largest drop: $\Delta = -0.011$ vs.\ A1, with Cohen's $q = 0.192$. MERT uses full fine-tuning with 95M parameters yet achieves lower correlation than MuQ's frozen 1M-parameter setup, reinforcing that encoder pre-training quality is the dominant factor. At the utterance level, the gap is more pronounced for MI (A4 SRCC $= 0.825$ vs.\ A1 SRCC $= 0.838$) than for TA (A4 SRCC $= 0.619$ vs.\ A1 SRCC $= 0.591$), suggesting that MERT's speech-oriented representations may partially compensate when text alignment is the target dimension. Despite its smaller size, MERT's three orders of magnitude more pre-training data (160K~hours vs.\ MuQ's 1.6K~hours) do not offset MuQ's architectural and training advantages for quality prediction.

\subsection{Degradation Concordance}

We apply four degradation types (MP3 compression, Gaussian noise, pitch shift, tempo stretch) at three severities to top-quartile clips ($N = 114$, fold~0). For each original-degraded pair, we compute the \emph{concordance rate}: the fraction of pairs where the model assigns a lower score to the degraded version. A concordance of 1.0 means the metric always detects the degradation; 0.5 is chance level. \tabref{tab:degradation} reveals that the metric reliably detects signal-level artifacts (MP3 at 32~kbps: 0.97; noise at SNR 10~dB: 0.99) but is insensitive to musical-structural distortions (pitch shift: $0.43$--$0.51$; tempo stretch: $0.46$--$0.58$), consistent with MuQ's pre-training normalizing pitch/tempo variation. The below-chance concordance for mild noise (0.39) suggests that low-level dithering may slightly increase predicted quality, possibly because the model associates subtle noise with production effects present in higher-rated training clips.

\begin{table}[t]
\centering
\caption{Degradation concordance (A3b, fold~0, $N = 114$). Chance level is 0.50.}
\label{tab:degradation}
\small
\begin{tabular}{@{}lccc@{}}
\toprule
Degradation & Severe & Moderate & Mild \\
\midrule
MP3 compression & 0.97 & 0.84 & 0.52 \\
Gaussian noise & 0.99 & 0.89 & 0.39 \\
Pitch shift & 0.51 & 0.43 & 0.43 \\
Tempo stretch & 0.58 & 0.49 & 0.46 \\
\bottomrule
\end{tabular}
\end{table}

\subsection{Data Efficiency}
\label{sec:data_efficiency}

\begin{figure}[t]
    \centering
    \includegraphics[alt={Line plot of utterance-level SRCC and PCC as a function of training set size for frozen and LoRA models showing LoRA sample efficiency advantage},width=\columnwidth]{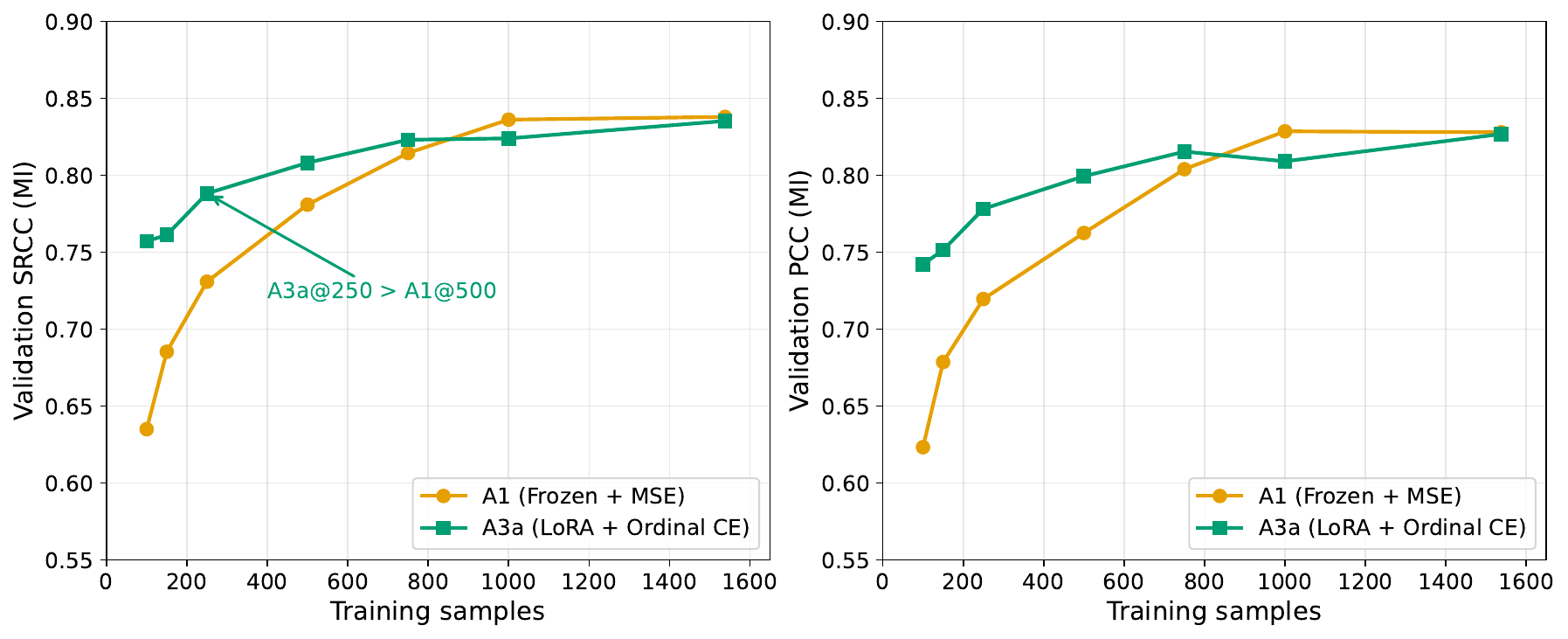}
    \caption{Data efficiency curves. A3a (LoRA) requires ${\sim}50\%$ fewer samples to reach comparable performance (crossover: A3a@250 $>$ A1@500).}
    \label{fig:data_efficiency}
\end{figure}

We train A1 and A3a on subsampled training sets (100--1,000 clips, fold~0). \tabref{tab:data_efficiency} shows A3a (LoRA) is consistently more sample-efficient: at $N = 100$, it achieves SRCC $= 0.757$ vs.\ A1's $0.635$. A crossover exists at A3a@250 ($0.788$) $>$ A1@500 ($0.781$), meaning LoRA halves the data requirement. With just 150 clips (comparable to one artist's discography), A3a achieves SRCC $= 0.761$, making personalized evaluators practical. Notably, both models exhibit diminishing returns beyond 1{,}000 clips, with the full training set (${\sim}1{,}540$) providing only marginal gains over the 1{,}000-clip subset. The convergence at the full data regime (A1: 0.838 vs.\ A3a: 0.835) suggests that LoRA's advantage is primarily in the low-data regime, where the encoder adaptation compensates for limited supervision. \figref{fig:data_efficiency} visualizes these curves.

\begin{table}[t]
\centering
\caption{Data efficiency: utterance-level MI correlation vs.\ training set size (fold~0).}
\label{tab:data_efficiency}
\small
\begin{tabular}{@{}rcccc@{}}
\toprule
\multirow{2}{*}{$N_\text{train}$} & \multicolumn{2}{c}{A1 (Frozen+MSE)} & \multicolumn{2}{c}{A3a (LoRA+Ord.)} \\
\cmidrule(lr){2-3} \cmidrule(lr){4-5}
 & SRCC & PCC & SRCC & PCC \\
\midrule
100 & 0.635 & 0.623 & 0.757 & 0.742 \\
150 & 0.685 & 0.679 & 0.761 & 0.751 \\
250 & 0.731 & 0.720 & 0.788 & 0.778 \\
500 & 0.781 & 0.763 & 0.808 & 0.799 \\
750 & 0.815 & 0.804 & 0.819 & 0.810 \\
1{,}000 & 0.836 & 0.829 & 0.824 & 0.809 \\
Full (${\sim}$1{,}540) & 0.838 & 0.828 & 0.835 & 0.827 \\
\bottomrule
\end{tabular}
\end{table}

\section{Discussion}
\label{sec:discussion}

\subsection{Why Do Frozen Features Suffice?}

The finding that frozen MuQ features with a simple MLP match more sophisticated recipes is consistent with LPIPS in vision~\cite{zhang2018lpips}. We hypothesize that MuQ's pre-training, which combines masked prediction and contrastive objectives on diverse music data, produces representations that already encode quality-relevant attributes such as spectral smoothness, harmonic structure, and timbral consistency. MOS prediction on MusicEval's wide quality range (spanning near-random noise to professional-quality outputs) appears to be linearly separable in this embedding space. Whether frozen features suffice for finer-grained discrimination (e.g., distinguishing between two high-quality systems that differ only subtly) remains an open question and a priority for future investigation.

\subsection{Personalized Evaluators}

The low data requirement opens the door to personalized quality evaluators. A listener who rates ${\sim}150$ familiar songs could train an evaluator achieving SRCC $> 0.76$. This enables an ``annotate one discography, evaluate all generations'' workflow: rate familiar music in a target style, then use the evaluator as a style-specific quality filter for generated outputs. This could serve niche creative communities, enable preference-aligned generation pipelines, and support A/B testing against personal quality standards.

Concretely, a content creator working in a specific genre (e.g., lo-fi hip-hop, ambient electronic) could annotate a small set of reference tracks with personal quality ratings, then fine-tune a personalized head using LoRA adaptation in under 10~minutes on consumer hardware. The resulting evaluator would reflect that creator's aesthetic preferences rather than the aggregate expert consensus encoded in the base model.

\subsection{System-Level vs.\ Utterance-Level Gap}

All models show substantially higher system-level (SRCC $\sim$0.95) than utterance-level (SRCC $\sim$0.84) correlation. This gap reflects the noise reduction from averaging: system-level scores aggregate $\sim$88 clips per TTM model ($2748/31$), smoothing prediction errors. The utterance-level SRCC of 0.84 is on par with NISQA's per-utterance performance on clean speech~\cite{mittag2021nisqa}, suggesting comparable per-sample reliability.

\subsection{TA Head Performance}

The textual alignment (TA) head achieves substantially lower correlation (SRCC $\sim$0.60) than the musical impression (MI) head (SRCC $\sim$0.84). This gap likely reflects a fundamental architectural limitation: TA assessment requires comparing audio content against the text prompt, which our architecture cannot access, as it processes only the audio waveform. Text-conditioned architectures (e.g., using CLAP embeddings of the prompt) would be needed to meaningfully address text-audio alignment prediction.

\subsection{Implications for Metric Design}

Practitioners should invest in encoder selection rather than complex training objectives. The correlation difference from encoder choice (Cohen's $q = 0.192$) dwarfs any training recipe effect ($q < 0.09$), consistent with speech quality prediction findings~\cite{ragano2024scoreq}. The system-level vs.\ utterance-level gap (SRCC 0.95 vs.\ 0.84) reflects noise reduction from averaging, with per-sample reliability comparable to NISQA on clean speech~\cite{mittag2021nisqa}. This finding suggests that for future music evaluation metrics, the community should prioritize improving encoder pre-training over developing increasingly sophisticated downstream training objectives.

\subsection{Computational Cost}

\ours{} is designed for accessibility. Inference takes approximately 35~ms per 10-second clip on an NVIDIA RTX~4080, with a peak VRAM footprint of 3~GB. Training the full pipeline (A1 baseline, 50 epochs, 5 folds) completes in under 2~GPU-hours on the same hardware, making it feasible for individual researchers without access to HPC clusters. The frozen-encoder models (A1, A2) require only ${\sim}$1M trainable parameters, and even the LoRA-adapted variant (A3a) adds only ${\sim}$2M parameters, orders of magnitude fewer than full fine-tuning of the 310M-parameter encoder.

\subsection{Limitations}

Several limitations warrant discussion. First, all results are on MusicEval with 31 TTM systems; cross-dataset generalization is untested. While MusicEval covers a broad quality range and diverse system architectures, it represents a snapshot of the TTM landscape as of early 2025. The single-dataset setting means that the observed quality range is bounded by the systems in MusicEval; whether the metric extrapolates to systems substantially better or worse than those in the training set remains to be validated.

Second, the metric is insensitive to musical-structural distortions (pitch shift: concordance $0.43$--$0.51$; tempo stretch: $0.46$--$0.58$), limiting its utility for detecting such errors. This likely reflects MuQ's pre-training objective, which is invariant to pitch and tempo variation by design. Addressing this limitation may require augmentation-aware fine-tuning or multi-task training with explicit distortion detection objectives.

Third, data efficiency and degradation experiments use fold~0 only; trends may shift with different folds. Fourth, the TA head's low correlation (SRCC $\sim$0.60) reflects the lack of text conditioning in the architecture: without access to the input prompt, the model cannot assess whether the generated audio matches its textual description.

Finally, expert ratings inherently reflect a particular aesthetic perspective. While MusicEval uses trained musicologists, their judgments may not align with preferences of diverse listener populations across genres, cultures, and listening contexts. The personalized evaluator capability partially addresses this, but requires individual annotation effort.

\section{Conclusion}
\label{sec:conclusion}

We presented \ours{}, a per-sample quality metric for generated music. Our systematic investigation reveals that frozen representations from a music understanding encoder (MuQ-310M) with a simple MLP head achieve system-level SRCC $= 0.957$ with expert quality ratings, comparable to the closed-source DORA-MOS system and $4.8\times$ higher than Audiobox Aesthetics. A progressive ablation over training objectives, encoder adaptation, and multi-task strategies shows that additional complexity does not meaningfully improve over this simple baseline, with all component deltas below $\Delta = 0.01$ at system level. Encoder choice is the single most impactful design decision.

A data efficiency analysis further reveals that LoRA-adapted models are remarkably sample-efficient: only 150 training clips (comparable to a single artist's discography) yield utterance-level SRCC $= 0.761$, and 250 clips suffice to match the frozen baseline trained on 500. This low annotation barrier makes personalized quality evaluators practical: a single listener can rate one artist's song catalog and then use the resulting evaluator to score generated music in that style, enabling style-specific quality control without broad annotation campaigns.

A controlled degradation analysis reveals selective sensitivity: the metric reliably detects signal-level artifacts (MP3 compression concordance $= 0.97$, noise $= 0.99$ at severe levels) but is insensitive to musical-structural distortions (pitch shift $0.43$--$0.51$, tempo stretch $0.46$--$0.58$), consistent with MuQ's pre-training normalizing pitch and tempo variation.

\ours{} requires only 35~ms and 3~GB VRAM per clip, making it accessible to individual researchers without dedicated compute clusters.

Key future directions include: (1)~cross-dataset validation on SongEval~\cite{songeval2025} and emerging benchmarks to test generalization; (2)~improving sensitivity to structural distortions such as pitch and tempo errors, potentially through augmentation-aware training; (3)~text-conditioned architectures that jointly predict quality and alignment; and (4)~personalized evaluators trained from individual listener preferences to support style-specific generation workflows.

\section{AI Usage Statement}

In accordance with the ISMIR 2026 AI Usage Policy, we declare the following use of AI tools in this work. Large language models (OpenAI GPT-4o and Anthropic Claude) were used to assist with refining the written prose of the manuscript and for debugging portions of the training and evaluation code. All AI-assisted outputs were reviewed, verified, and edited by the authors. The experimental design, research methodology, data analysis, and scientific conclusions are entirely the work of the authors.

\section{Ethics Statement}

No human subjects research was conducted; all experiments use the publicly available MusicEval dataset under its original license. We acknowledge that quality assessment is inherently subjective and that MusicEval's annotations reflect one group of expert annotators whose judgments may not generalize across all musical traditions. The personalized evaluator capability could reinforce aesthetic biases if deployed without awareness of its training distribution.

\bibliography{references}

\end{document}